%% file: main.tex
\documentclass[letterpaper, 10 pt, conference]{ieeeconf}
\IEEEoverridecommandlockouts

\usepackage[T1]{fontenc}
\usepackage{amsmath,amsfonts}
\usepackage{array}
\usepackage{cite}
\usepackage[caption=false,font=scriptsize]{subfig}
\usepackage{textcomp}
\usepackage{stfloats}
\usepackage{url}
\usepackage{verbatim}
\usepackage{graphicx}
\usepackage{amssymb}
\usepackage{pifont}
\usepackage[skip=2pt]{caption}
\usepackage[hidelinks]{hyperref}
\usepackage[none]{hyphenat}
\usepackage[capitalize]{cleveref}
\usepackage{comment}
\usepackage{multirow}
\usepackage{booktabs}
\usepackage{rotating}
\usepackage{makecell}
\usepackage{tabularx}
\usepackage{xcolor}
\usepackage[acronym]{glossaries}
\makeglossaries
\newacronym{llm}{LLM}{Large Language Model}
\newacronym{vlm}{VLM}{Vision Language Model}
\newacronym{ads}{ADS}{Automated Driving Systems}
\newacronym{ttc}{TTC}{Time-to-Collision}
\newacronym{mdc}{MDC}{Minimum Distance to Collision }
\newacronym{ev}{EV}{Ego vehicle}

\usepackage[linesnumbered]{algorithm2e}
\RestyleAlgo{ruled}
\DontPrintSemicolon 

\usepackage{blindtext}
\usepackage{siunitx}
\AtBeginDocument{\RenewCommandCopy\qty\SI}
\usepackage{tikz}
\usetikzlibrary{arrows.meta}
\usetikzlibrary{arrows}
\usepackage{pgfplots}
\pgfplotsset{compat=1.18}
\usepackage{xcolor}
\definecolor{Bluelight}{HTML}{0065BD} % TUMBlue
\definecolor{Black}{HTML}{000000}
\definecolor{Blue}{HTML}{005293}
\definecolor{Bluestrong}{HTML}{003359}
\definecolor{Red}{HTML}{8C000F}
\definecolor{Grey}{HTML}{808080}
\definecolor{Greylight}{HTML}{CCCCCC}
\definecolor{Orange}{HTML}{E37222}
\definecolor{Green}{HTML}{A2AD00}
\usepackage[absolute,overlay]{textpos}

\def\BibTeX{{\rm B\kern-.05em{\sc i\kern-.025em b}\kern-.08em
    T\kern-.1667em\lower.7ex\hbox{E}\kern-.125emX}}

\usepackage{caption}
\begin{document}

%\title{\LARGE \bf LLM Behind the Wheel: Evaluating and Generating Safety-Critical Driving Scenarios}

\title{\LARGE \bf From Words to Collisions: LLM-Guided Evaluation and Adversarial Generation of Safety-Critical Driving Scenarios}

\author{Yuan Gao$^{1}$, Mattia Piccinini$^{1}$, Korbinian Moller$^{1}$, Amr Alanwar$^{2}$,  Johannes Betz$^{1}$  % <-this % stops a space
%\thanks{Manuscript received XXX, 2022; revised XXX 2022. \textit{(Corresponding author: Tobias Betz (email: tobi.betz@tum.de)}}
\thanks{$^{1}$ Y. Gao, M. Piccinini, K. Moller and J. Betz are with the Professorship of Autonomous Vehicle Systems, TUM School of Engineering and Design, Technical University of Munich, 85748 Garching, Germany; Munich Institute of Robotics and Machine Intelligence (MIRMI), \{yuan\_avs.gao, mattia.piccinini, korbinian.moller, johannes.betz\}@tum.de}
\thanks{$^{2}$ A. Alanwar is with  the TUM School of Computation,
Information and Technology, Department of Computer Engineering,
Technical University of Munich, 74076 Heilbronn, Germany. (e-mail: alanwar@tum.de)}
}% <-this % stops a space

% The paper headers

% Remember, if you use this, you must call \IEEEpubidadjcol in the second
% column for its text to clear the IEEEpubid mark.

\maketitle
%\thispagestyle{empty} %--to make title page number less
%\pagestyle{empty}    % -- to make other pages number less.
%

%\input{chapters/0_abstract}

%%%%%%%%%%%%%%%%%%%%%%%%%%%%%%%%%%%%%%%%%%%%%%%%%%%%%%%%%
%%% Abstract
%%%%%%%%%%%%%%%%%%%%%%%%%%%%%%%%%%%%%%%%%%%%%%%%%%%%%%%%%

\begin{abstract}
Ensuring the safety of autonomous vehicles requires virtual scenario-based testing, which depends on the robust evaluation and generation of safety-critical scenarios. 
So far, researchers have used scenario-based testing frameworks that rely heavily on handcrafted scenarios as safety metrics. To reduce the effort of human interpretation and overcome the limited scalability of these approaches, we combine \glspl{llm} with structured scenario parsing and prompt engineering to automatically evaluate and generate safety-critical driving scenarios. We introduce Cartesian and Ego-centric prompt strategies for scenario evaluation, and an adversarial generation module that modifies trajectories of risk-inducing vehicles (ego-attackers) to create critical scenarios. We validate our approach using a 2D simulation framework and multiple pre-trained \glspl{llm}. The results show that the evaluation module effectively detects collision scenarios and infers scenario safety. Meanwhile, the new generation module identifies high-risk agents and synthesizes realistic, safety-critical scenarios. We conclude that an LLM equipped with domain-informed prompting techniques can effectively evaluate and generate safety-critical driving scenarios, reducing dependence on handcrafted metrics. %Future studies will investigate using 3D simulation environments and LLM fine-tuning to improve the quality of the scenarios. 
We release our open-source code and scenarios at: \href{https://github.com/TUM-AVS/From-Words-to-Collisions}{https://github.com/TUM-AVS/From-Words-to-Collisions}.
\end{abstract}

%%%%%%%%%%%%%%%%%%%%%%%%%%%%%%%%%%%%%%%%%%%%%%%%%%%%%%%%%
%%% Keywords
%%%%%%%%%%%%%%%%%%%%%%%%%%%%%%%%%%%%%%%%%%%%%%%%%%%%%%%%%


%%%%%%%%%%%%%%%%%%%%%%%%%%%%%%%%%%%%%%%%%%%%%%%%%%%%%%%%%
%%% Introduction
%%%%%%%%%%%%%%%%%%%%%%%%%%%%%%%%%%%%%%%%%%%%%%%%%%%%%%%%%

\section{Introduction}
\label{sec:introduction}
The development and deployment of autonomous vehicles have progressed rapidly, reducing human intervention within specific Operational Design Domains (ODDs) step by step. Companies like Waymo have deployed fully autonomous SAE Level 4~\cite{Betz2024} robotaxi services in defined ODDs, demonstrating the potential of driverless technology in urban environments. This advancement is primarily driven by developing and validating highly reliable \gls{ads}. Traditionally, validation has relied on real-world testing, including on-road trials~\cite{khan2023safety}. However, real-world testing cannot fully capture the diversity of driving situations and edge cases. To address this, researchers and industry have increasingly employed virtual scenario-based testing, enabling cost-effective simulation of realistic conditions~\cite{riedmaier2020survey}. %and targeted design of rare, safety-critical cases often missing from real-world~\cite{menzel2018scenarios}.

\begin{figure}[t]
    \centering
    \includegraphics[width=0.495\textwidth]{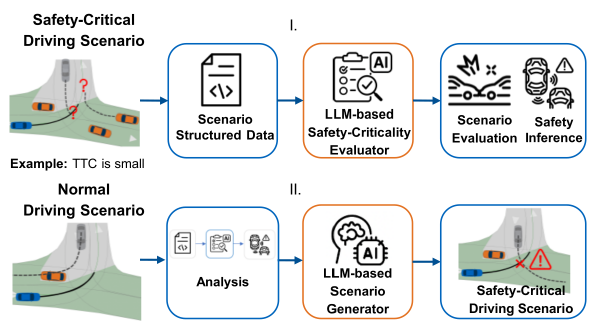}
    \caption{Our paper presents a new LLM-based framework with two modules: Module I: LLM-based evaluation of safety-critical driving scenarios. Module II: LLM-based generation of safety-critical driving scenarios.}
    \label{fig: concept}
\end{figure}
Scenario-based testing needs to evaluate the behavior of \gls{ads} in safety-critical contexts. Various metrics have been proposed to assess the safety-criticality of driving scenarios~\cite{hallerbach2018simulation}, including proximity-based indicators like \gls{ttc} and \gls{mdc}, categorized into temporal and non-temporal types. Based on these metrics, test engineers can evaluate the safety-criticality of a scenario. 

Since the release of GPT-3~\cite{brown2020language}, applications using  Large
Language Models (LLMs) have surged. As large-scale pre-trained models on general-purpose datasets~\cite{vaswani2017attention}, \glspl{llm} combine strong generalization with adaptability, allowing direct application to domain-specific tasks via prompting or fine-tuning~\cite{sahoo2024systematic}. Recently, researchers have explored their use in autonomous driving tasks, including perception~\cite{wu2023language}, motion planning~\cite{wang2024dualad}, and vehicle control~\cite{xu2024drivegpt4}. This raises a key question: \textit{Given their good generalization and adaptability, can LLMs be used to evaluate and generate safety-critical driving scenarios, as illustrated in Fig.~\ref{fig: concept}?}

\section{Related Work}
Evaluating and generating safety-critical scenarios are key to scenario-based testing in autonomous driving. This section reviews both traditional and \glspl{llm}-based approaches.
\subsection{Safety-Critical Driving Scenario Evaluation for \gls{ads}}
Several recent works have explored evaluating safety-critical driving scenarios using different methods. In~\cite{song2023critical}, safety-critical scenarios were defined as collisions or near-collisions with the ego vehicle. In~\cite{hallerbach2018simulation}, a set of safety metrics was employed in scenario evaluation, including \gls{ttc}, braking behavior, required deceleration, and traffic-related indicators. While these safety metrics can be computed from scenario data, analyzing them to assess safety requires human effort and domain knowledge.

\textbf{\glspl{llm} as evaluator}: Recent studies have explored \glspl{llm} for driving scenario evaluation, focusing on realism, behavior, and consistency.  In~\cite{wu2024reality}, they prompted \glspl{llm} with contextual data, e.g., road layout, weather, vehicle behavior, to assess scenario realism using the DeepScenario dataset~\cite{lu2023deepscenario}. \cite{you2025comprehensive} proposes a framework incorporating high-level evaluation criteria, including performance, safety, and comfort, alongside scenario data from the CARLA~\cite{Dosovitskiy17}  to evaluate driving style and performance. However, in~\cite{you2025comprehensive} they focus on assessing final driving style and driving performance level based on aggregated input, rather than directly assessing individual metrics like safety. 
OmniTester~\cite{lu2024multimodal} extended text-only evaluation to multimodal analysis by generating captions from visualized scenarios using a vision-language model and comparing them with original descriptions from \gls{llm}. Notably, a recent survey~\cite{gao2025foundation} summarizes existing \gls{llm}-based scenario analysis methods. However, none of the papers explicitly assess safety criticality in driving scenarios by using \glspl{llm}.

%While these studies demonstrate the applicability of \glspl{llm} in evaluating realism, behavior, and consistency in driving scenarios, none have investigated their ability to assess the safety criticality of driving scenarios.

% While effective in evaluating plausibility and behavior, 
\subsection{Safety-Critical Driving Scenario Generation for \gls{ads}}
Many studies have explored the generation of safety-critical scenarios to test \gls{ads}. According to the survey \cite{ding2023survey}, the main approaches are data-driven, adversarial, and knowledge-based.
Data-driven methods use real-world datasets to generate realistic scenarios and avoid overfitting, leveraging models like Bayesian networks~\cite{wheeler2016factor} and deep generative models~\cite{ding2018new}.
Adversarial approaches aim to expose weaknesses in \glspl{ads} by synthesizing rare but high-risk situations using techniques such as differentiable renderers~\cite{jain2019analyzing} or reinforcement learning~\cite{sun2021corner}.
Knowledge-based methods integrate domain expertise, e.g., traffic rules, through manual rules~\cite{rana2021building} or hybrid approaches combining rule-based priors with adversarial policy learning~\cite{cao2023robust} for enhanced controllability.

\textbf{\glspl{llm} as generator}: Recently, \glspl{llm} have emerged as powerful tools for generating simulation scenarios directly from natural language.  Depending on the abstraction level and application, the underlying simulators fall into two categories: microscopic, such as CARLA~\cite{Dosovitskiy17}, focusing on ego vehicle behavior, and macroscopic, like SUMO~\cite{SUMO2018}, enabling large-scale traffic simulations. At the microscopic level, ChatScene~\cite{zhang2024chatscene} and TTSG~\cite{ruan2024traffic} used \glspl{llm} to generate safety-critical scenarios in CARLA from textual descriptions. ChatScene further utilized these scenarios for training and evaluating vehicle control algorithms, while TTSG applied them to multi-agent planning. For macroscopic scenario generation, ChatSUMO~\cite{li2024chatsumo} and OmniTester~\cite{lu2024multimodal} leveraged \glspl{llm} to produce realistic urban traffic simulations from text, showcasing the scalability of language-driven scenario generation.
\subsection{Critical Summary}
To the best of our knowledge, the existing literature is limited by at least one of the following aspects:
\begin{enumerate}
    \item The safety-criticality of scenarios is typically evaluated by test engineers using predefined metrics. For example, ChatScene~\cite{zhang2024chatscene} and TTSG~\cite{ruan2024traffic} computed metrics such as \gls{ttc} and \gls{mdc}, but then a human had to interpret the generated scenarios.
    
    \item No examples of \glspl{llm} to evaluate safety-critical scenarios.
    %Limited research explored \glspl{llm} for scenario evaluation. 
    The authors of \cite{wu2024reality} and~\cite{lu2024multimodal} analyzed the scenario realism and semantic consistency, while~\cite{Dosovitskiy17} evaluated the overall driving style and performance. However, no one has performed any safety-criticality assessment so far.
    
    \item Existing LLM-based scenario generators relied on textual descriptions provided by experts. Integrating adversarial methods into \glspl{llm} to generate safety-critical scenarios remains unexplored. %It has the potential to deliver greater efficiency and controllability. 
    %Compared to the purely text-based approaches of OmniTester~\cite{lu2024multimodal} and ChatSUMO~\cite{li2024chatsumo}, we use \glspl{llm} to identify an ego-attacker vehicle and change its trajectory on purpose.
\end{enumerate}

\subsection{Contributions}
To address the previous limitations, the key contributions of this paper are the following:

\begin{itemize} 
    \item We present a novel framework that integrates \glspl{llm} as evaluation modules to assess the safety-criticality of structured driving scenarios.
    
    \item We compare different prompt formulations to analyze safety-critical scenarios and focus on two use cases: collision evaluation and safety inference. %to evaluate recorded simulation scenarios and to perform risk inference for individual traffic participants. 

    \item We propose a novel \gls{llm}-based module that transforms safe scenarios into safety-critical ones by adversarially modifying the trajectories of selected agents.
    %We devise a novel \gls{llm}-based module that generates safety-critical scenarios starting from safe scenarios, by adversarially modifying specific agents' trajectories to create safety-critical cases.

    \item We present empirical evaluations through ablation studies on different prompt formulations and multiple state-of-the-art \glspl{llm}, tested on 200 randomly selected scenarios from a dataset of 6000 simulations.
\end{itemize}

\begin{figure*}[t]
    \centering
    \includegraphics[width=1\textwidth]{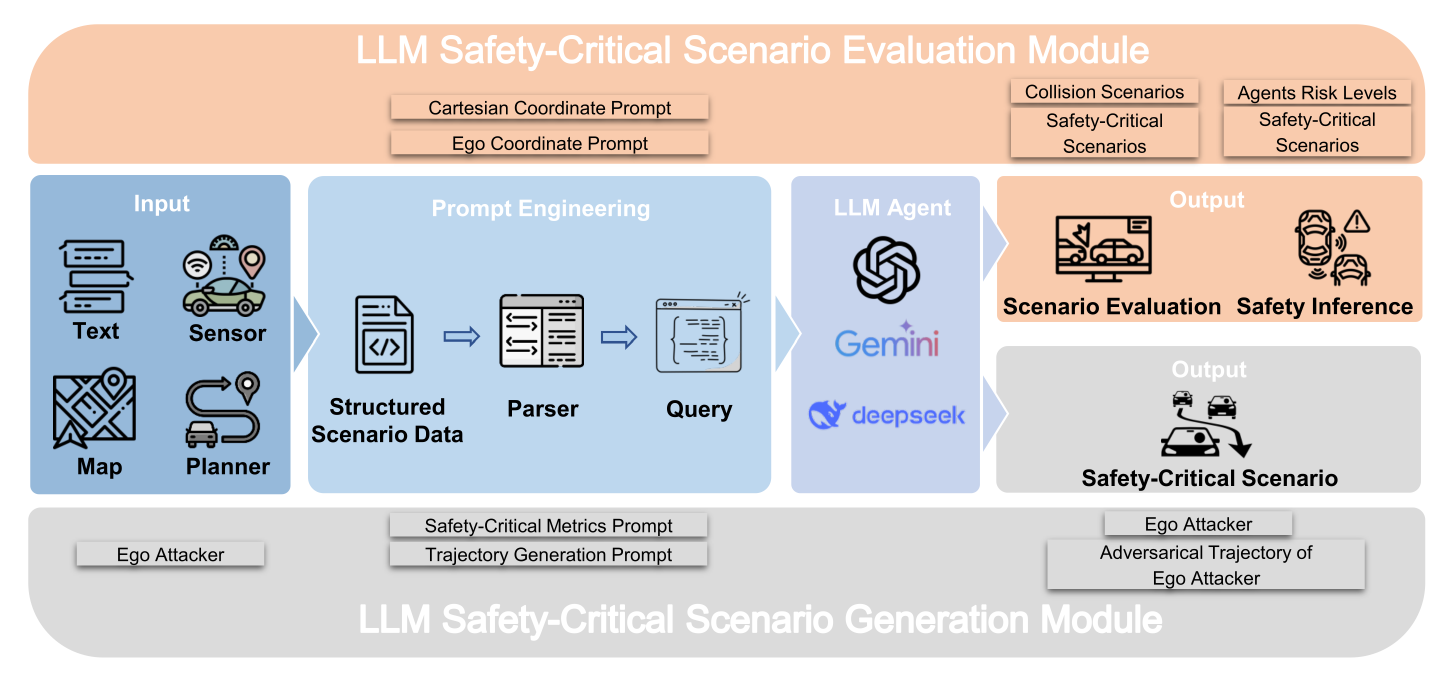} 
    \caption{Overview of the proposed framework based on \gls{llm} to evaluate and generate safety-critical scenarios.}
    \label{fig: framework} % Optional: for referencing the figure
\end{figure*}
%%%%%%%%%%%%%%%%%%%%%%%%%%%%%%%%%%%%%%%%%%%%%%%%%%%%%%%%%
%%% Methodology
%%%%%%%%%%%%%%%%%%%%%%%%%%%%%%%%%%%%%%%%%%%%%%%%%%%%%%%%%
\section{Methodology}
\label{sec:method}

This paper proposes a novel \gls{llm}-based framework (Fig.~\ref{fig: framework}) for evaluating and generating safety-critical driving scenarios.
Primarily, we focus on scenarios from a 2D simulator like CommonRoad\cite{althoff2017commonroad}, which supports scenario-based tests for the motion planning algorithms.
 
 % It is noted that all of the scenarios are recorded from the real world, and simulation scenarios consist of structured datasets that include detailed information about map information, road category, traffic signs, static and dynamic traffic participants, and a motion planning task defined for the ego vehicle.

\textbf{Module I-Evaluation} (top parts of Fig. \ref{fig: concept}--\ref{fig: framework}): In the \gls{llm}-based scenario evaluation module, a key challenge is that \glspl{llm} require natural language input, making direct processing of structured scenario data difficult. To address this, we introduce a parser converting structured data into natural language descriptions, serving as contextual input for the \gls{llm}. To enhance domain-specific adaptability for scenario evaluation and safety inference, we propose and compare Cartesian and Ego coordinate prompt templates.

\textbf{Module II-Generation} (bottom parts of Fig. \ref{fig: concept}--\ref{fig: framework}): We propose an \gls{llm}-based generation module that transforms safe scenarios into safety-critical ones. Our framework identifies potential ego-attackers by analyzing the agents' motion and safety-critical metrics (\gls{ttc} and \gls{mdc}), provided through prompts. Then, our model modifies the ego-attackers' trajectories adversarially to induce collisions with the \gls{ev}.

\subsection{Structured Driving Scenario Data}
This section describes the driving scenario data used as input by our framework ("Input" and "Structured Scenario Data" blocks in Fig. \ref{fig: framework}).
We employ a 2D simulator providing a standardized scenario representation, which contains information on the road map, road network, traffic signs, traffic participants, and motion planning task. This structure supports the testing and benchmarking motion planning algorithms across diverse driving scenarios.

\begin{figure}[htbp]
    \centering
    \includegraphics[width=0.9\linewidth]{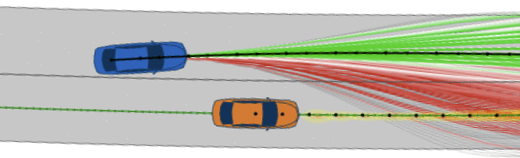}
    \caption[Example of an overtaking scenario]%
    {Example of an overtaking scenario in the 2D simulator with the motion planner used in this paper\textsuperscript{1}.}
    \label{fig:commonroadfrenetix}
\end{figure}

In our scenario representation, the road network uses the lanelet map~\cite{bender2014lanelets}, which is aligned with OpenStreetMap (OSM) standards, and includes detailed lane boundaries, connectivity, and traffic flow topology. The driving scenario incorporates traffic infrastructure elements (e.g., signs, traffic lights) and various traffic participants, such as cars, buses, and motorcycles, which are classified as static or dynamic obstacles. All entities are annotated with geometric and kinematic attributes to simulate real-world interactions accurately. The scenario data also includes the inputs for a motion planning problem, which specifies the \gls{ev}’s initial state, goal states, and the planning task (Fig. \ref{fig:commonroadfrenetix}).
\vspace{-1.1 em}
\vspace{-1.1 em}
\addtocounter{footnote}{1}
\footnotetext[\value{footnote}]{We use the CommonRoad simulator~\cite{althoff2017commonroad} and the Frenetix motion planner~\cite{Frenetix}, but other 2D simulators and motion planner could be equivalently used.}

To generate a dataset of scenarios for our framework, we perform a set of multi-vehicle open-loop simulations, in which the \gls{ev}'s motion planner tries to generate collision-free trajectories to reach a planning goal. We collect these simulation scenarios in a structured dataset, which will be used as input by our framework. We define our structured scenario data as:
\begin{equation}
   S = \{\mathcal{L}, \mathcal{T}, \mathcal{O}, \mathcal{E}\}
\end{equation}
where $\mathcal{L}$ represents the lanelets, $\mathcal{T}$ denotes the traffic signs, $\mathcal{O}$ is the set of obstacles, and $\mathcal{E}$ represents the \gls{ev} state.  

\subsection{\gls{llm}-based Safety-Critical Scenario Evaluation Module}
\subsubsection{Parser}
To bridge the gap between the structured scenario dataset $\mathcal{S}$ and the natural language input required by the \gls{llm}, we introduce a parser that transforms structured scenario data into textual descriptions (Fig. \ref{fig: framework}). Specifically, we define a parsing function $\mathcal{F}$ that maps scenario data $\mathcal{S}$ to a context $\mathcal{C}$, such that $\mathcal{C} = \mathcal{F}(\mathcal{S})$.

%
% \begin{equation}
%     \mathcal{C} = \mathcal{F}(\mathcal{S})
% \end{equation}
%
This parsing function extracts key elements from $\mathcal{S}$, including lanelets information $\mathcal{L}$, obstacles states $\mathcal{O}$, and \gls{ev} states $\mathcal{E}$, excluding traffic signs $\mathcal{T}$, which are not required for evaluation. Obstacles and \gls{ev} data are represented in global Cartesian coordinates, including position, velocity, orientation, and acceleration. 

\begin{figure}[t]
    \centering
    \includegraphics[width=0.475\textwidth]{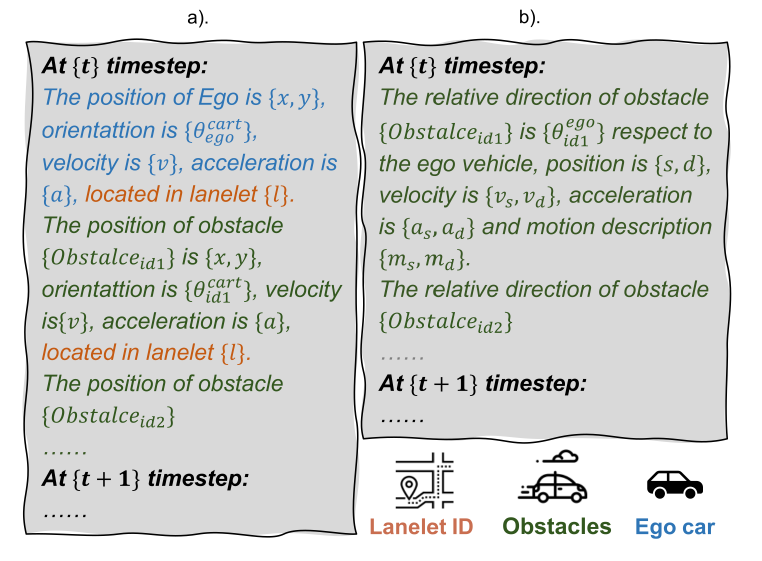}
    \caption{Output from the Cartesian-coordinate parser (a) and Ego-centric parser (b). The blue part describes the ego vehicle's state, the orange parts are the lanelet's identification numbers, and the green part describes the obstacles' states.}
    \label{fig: Parser_analysis}
\end{figure}

%To support different perspectives in our scenario evaluation module, 
We design two specialized parsers: one using a map-based representation in Cartesian coordinates, named $\mathcal{F}_{\text{cart}}$, and the other using an ego-centric description, named $\mathcal{F}_{\text{ego}}$.
$\mathcal{F}_{\text{cart}}$ describes the absolute states of all agents in Cartesian coordinates, and its output is illustrated in Fig.~\ref{fig: Parser_analysis} (a)
%We design a Cartesian-specific parser function, $\mathcal{F}_{\text{cart}}$, which captures the absolute states of all agents for evaluating safety.
In contrast, the Ego-centric parser $\mathcal{F}_{\text{ego}}$ describes the scenario from the \gls{ev}'s perspective, encoding relative spatial relations (e.g., front-rear, left-right) and dynamics (e.g., position, velocity, acceleration) using the longitudinal and lateral coordinates $\{s,d\}$. It also generates relative motion descriptions for each obstacle, such as
“$m_s$: The obstacle is approaching the \gls{ev} longitudinally from the front”, 
“$m_d$: The obstacle is moving laterally towards the \gls{ev} from the left”. The output of $\mathcal{F}_{\text{ego}}$ is shown in Fig.~\ref{fig: Parser_analysis}(b). More details about these motion descriptions can be found in our open-source code repository. 
The outputs of $\mathcal{F}_{\text{cart}}$ and $\mathcal{F}_{\text{ego}}$ provide the context for the following Cartesian and Ego prompts, described in the next subsection.

\subsubsection{Prompt Formulation} \label{sec:prompt_formul}

Since \glspl{llm} are pre-trained on vast amounts of unlabeled data, techniques like prompt engineering or fine-tuning~\cite{sahoo2024systematic} are commonly used to adapt the \glspl{llm} to downstream tasks. Given the strong reasoning capabilities of state-of-the-art models such as GPT-4o~\cite{achiam2023gpt}, prompt-based adaptation is often favored over fine-tuning. In this prompt approach, a system message defines the model’s behavior and domain, while a user message specifies task instructions, enabling adaptation without changing the model's weights. Specifically, we use prompting to adapt \glspl{llm} to evaluate safety-critical scenarios.

We propose a structured prompting template integrating advanced techniques with Contextual Prompting, Chain-of-Thought reasoning, and In-Context Learning. Within this template, we define two prompt types based on the contextual input: the Cartesian coordinate prompt ($\textit{Prompt}_{\mathrm{cart}}$) and the Ego coordinate prompt ($\textit{Prompt}_{\mathrm{ego}}$). 
In both cases, a system message assigns the \gls{llm} the role of a collision evaluation expert, enabling reasoning over safety-critical metrics such as \gls{ttc} and \gls{mdc}. The key difference lies in the user message, which considers the scenario information based on Cartesian or Ego coordinates (Fig.~\ref{fig: Parser_analysis}). In Section \ref{sec:results}, we compare the performance of the two prompts. The following paragraphs detail the techniques that we employ in our prompt design.

\textbf{Contextual Prompting (CP)}:
The system message defines key metrics, such as \gls{ttc} and \gls{mdc}, and quantifies safety-criticality by assigning risk scores based on threshold values for these metrics. By embedding this domain-specific knowledge, CP enhances the model’s understanding of evaluation criteria, improving the reliability of its responses. Furthermore, the framework remains flexible, allowing additional metrics to be easily incorporated.

\textbf{Chain-of-Thought (CoT)}:
CoT prompting explicitly guides the \gls{llm} to reason step-by-step rather than providing immediate answers. In our framework, the user message requests intermediate reasoning steps such as assessing obstacles' trajectories, identifying critical obstacles, and evaluating the scenario's safety-criticality.

\textbf{In-Context Learning (ICL)}:
This prompt incorporates examples of collision and risk assessments within the system message. These demonstrations guide the \gls{llm}'s reasoning process, define the task format and evaluation logic.

By integrating the CP, CoT, and ICL techniques within $\mathrm{\textit{Prompt}_{\mathrm{cart}}}$ and $\mathrm{\textit{Prompt}_{\mathrm{ego}}}$, we adapt a general-purpose \gls{llm} to a specialized evaluator of driving scenarios. In Section \ref{sec:results}, ablation studies will assess how these techniques impact the model’s performance.

\subsection{\gls{llm}-Based Safety-Critical Scenario Generation Module}
We extend our framework with an adversarial scenario generation module that transforms safe driving scenarios into safety-critical ones. Our method identifies potential ego-attackers and adversarially changes their trajectories, based on agent-level risk scores.
\subsubsection{Parser}
To identify ego-attackers, we directly integrate the values of the \gls{ttc} and \gls{mdc} safety-critical metrics into the evaluation prompt. For this purpose, we design a safety-critical metrics parser that converts structured metric data into natural language context for the \gls{llm}, as shown in Fig.~\ref{fig: Parser_generation}(a).
Once an ego-attacker is identified, we use a trajectory generation parser to convert its recorded behavior into textual format, illustrated in Fig.~\ref {fig: Parser_generation}(b).

\subsubsection{Prompt Formulations}
\label{sec:prompt_generation}
In the generation of safety-critical scenarios, we combine two prompts: a safety-critical metrics prompt, to assess the obstacles' risk scores, and a trajectory generation prompt, to synthesize new adversarial trajectories for the ego attackers.

\begin{figure}[t]
    \centering
    \includegraphics[width=0.495\textwidth]{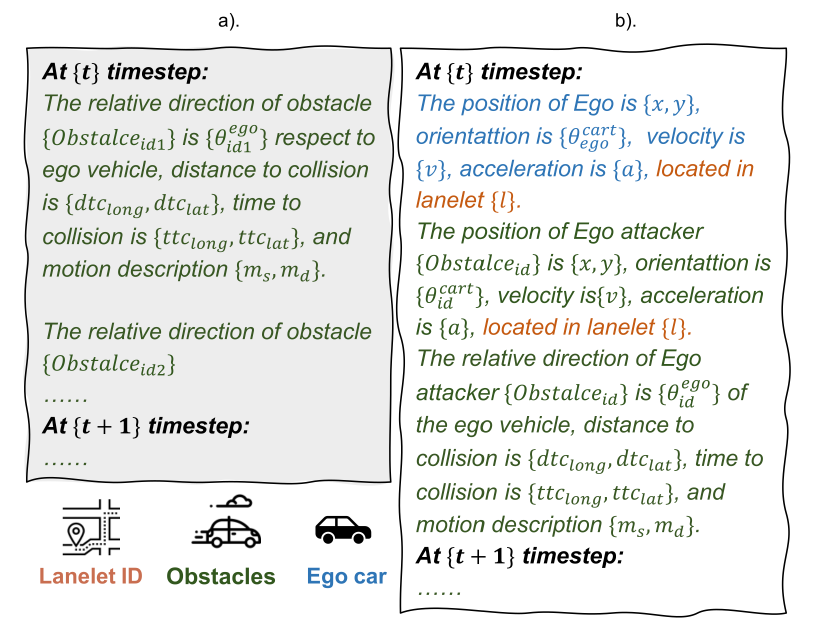}
    \caption{Output from safety-critical metrics parser (a) and trajectory generation parser (b). The blue part describes the ego state, the orange parts are the lanelet identification numbers, and the green parts are the obstacles' states.}
    \label{fig: Parser_generation}
\end{figure}

The safety-critical metrics prompt follows the structure of $\textit{Prompt}_{\mathrm{cart}}$ and $\textit{Prompt}_{\mathrm{ego}}$, with the \gls{llm} acting as a domain expert in collision evaluation. It incorporates CoT reasoning and ICL examples to guide the assessment of overall risk scores for each obstacle, based on thresholds for each safety-critical metric. Our prompt also includes motion descriptions, as shown in Fig.~\ref{fig: Parser_generation}(a), focusing on obstacles approaching from the front or sides, since these directions are  relevant to evaluate the planner's performance. This structure enables the \gls{llm} to reason about ego-attackers using both quantitative metrics and spatial context.

For trajectory generation, we adopt a different prompting strategy as the \gls{llm} transitions from evaluator to generator. The system message instructs the model to synthesize an adversarial trajectory that increases the collision likelihood with the \gls{ev}, given the scenario context.
To enhance the generation performance, we apply Self-Consistency (SC) by prompting the model to produce multiple adversarial trajectory candidates for the same ego-attacker.
Each candidate is scored using the safety-critical metrics prompt, and the one with a low-risk score for the ego-attacker is selected.
\subsection{Structured Output Format} \label{sec:output}
In our \gls{llm}-based scenario evaluation and generation framework, we enforce structured output formats using predefined schemes to ensure consistency and reduce hallucination~\cite{jiang2023structgpt}. These formats define fixed keys aligned with specific use cases, including collision evaluation, safety-critical scenario inference, agent-level risk assessment (e.g., risk scores), and trajectory generation parameters (e.g., position). The structure remains flexible and can be adapted to various prompt designs and task requirements.
\section{Results \& Discussion}
\label{sec:results}
% To recall the proposed research question:  \textit{could \glspl{llm} be leveraged effectively to analyze safety-critical aspects of autonomous driving scenarios and even be used to generate safety-critical scenarios?}
%This section evaluates the performance of our \gls{llm}-based framework across three use cases:
%
%\begin{itemize}
%\item Collision evaluation: detecting whether a collision occurs for the \gls{ev} in a driving scenario;
%\item Safety inference: predicting whether a scenario is safety-critical for the \gls{ev};
% \%item Generation of safety-critical scenarios.
%\end{itemize}

%scenario-level safety evaluation of recorded scenarios, online risk evaluation for the ego vehicle, and safety-critical scenario generation.

\subsection{Experimental Setup}
\label{sec:setup}
We conduct our experiments using the 2D simulator CommonRoad\cite{althoff2017commonroad}, with the Frenetix~\cite {Frenetix} open-source motion planner. This simulation setup enables us to collect diverse driving scenario datasets to be used as input for our framework. In our experiments, we compare the latest \glspl{llm} Gemini-1.5Pro~\cite{team2023gemini}, GPT4o, and DeepSeek-V3~\cite{bi2024deepseek}, which are the top powerful \glspl {llm} in the open leaderboard, using their APIs.
%  according to the actual ranking. 

All experiments are performed on a Dell Alienware R15 equipped with an Intel i7-13700KF CPU, a NVIDIA RTX 4090 GPU, and a 64 GB RAM. We simulate 6000 real recorded driving scenarios across different countries' maps, road categories, obstacle classes, and number of traffic participants.
The Frenetix motion planner successfully generates collision-free trajectories in over 4700 scenarios, allowing the \gls{ev} to reach its goal state. In approximately 1000 scenarios, the \gls{ev} collides with a traffic participant. The remaining scenarios involve other failures or exceptional conditions, such as unsolvable cases.
We use the collision cases to evaluate our scenario evaluator, treating them as ground truth for the collision evaluation use case. For the safety inference use case, we further identify safety-critical obstacles by computing \gls{ttc} and \gls{mdc} with risk score as ground truth. Lastly, for the scenario generation use case, we assess our generator by transforming safe scenarios into safety-critical ones through adversarial trajectory modifications.
\begin{figure}[t]
    \centering
    \includegraphics[width=0.495\textwidth]{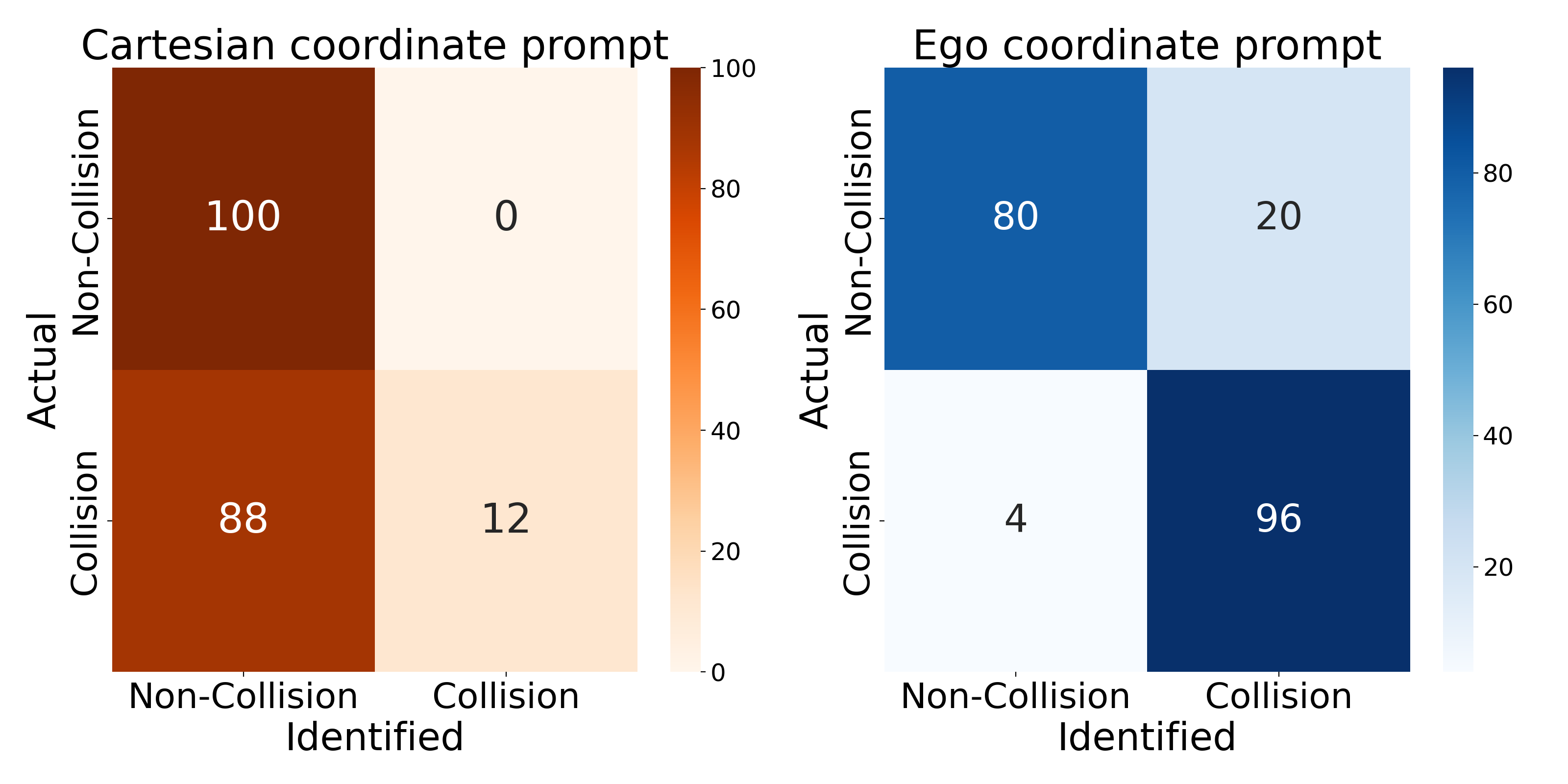}
    \caption{Confusion matrices for collision evaluation using $\mathrm{\textit{Prompt}_{\mathrm{cart}}}$ and $\mathrm{\textit{Prompt}_{\mathrm{ego}}}$.}
    \label{fig: confusion_scenarios}
\end{figure}

\subsection{Collision Evaluation in Driving Scenarios}
This use case evaluates if collisions occur with the \gls{ev} in driving scenarios. The inputs of our \gls{llm} include a map with geometric lanelet information, dynamic states of the \gls{ev}, and of all traffic participants throughout the simulation's duration. Our model outputs a boolean flag to indicate if a collision occurs for the \gls{ev} in the given scenario. We randomly select 100 non-collision and 100 collision scenarios for evaluation from the 6000 total scenarios.

To test our \gls{llm}-based collision evaluator, we employ the two prompt types $\mathrm{\textit{Prompt}_{\mathrm{cart}}}$ and $\mathrm{\textit{Prompt}_{\mathrm{ego}}}$ [Section~\ref{sec:prompt_formul}] with collision evaluation output format [Section~ \ref{sec:output}], using the Gemini-1.5Pro model. The results are summarized in Fig.~\ref{fig: confusion_scenarios}, which presents the confusion matrices for each prompt setting. $\mathrm{\textit{Prompt}_{\mathrm{cart}}}$ fails to detect most collisions, identifying only 12\% of collision scenarios correctly and resulting in an 88\% false negative rate. This indicates that Cartesian-based input is insufficient for accurate scenario risk assessment. In contrast, $\mathrm{\textit{Prompt}_{\mathrm{ego}}}$ achieves 96\% accuracy on collisions and 80\% on non-collisions, demonstrating the effectiveness of ego-centric inputs for safety-critical evaluation from the \gls{ev}'s perspective.

\subsubsection{Ablation Study} \label{sec:ablation}
The performance of our \gls{llm}-based evaluator depends on prompt design and advanced techniques with CP, ICL, and CoT reasoning, as described in Section~\ref{sec:prompt_formul}. To evaluate their impact, an ablation study is conducted for both $\mathrm{\textit{Prompt}_{\mathrm{cart}}}$ and $\mathrm{\textit{Prompt}_{\mathrm{ego}}}$: \textit{Base} (no techniques), \textit{CP} (CP only), \textit{CoT} (CP + CoT), and \textit{All} (CP + CoT + ICL).

\begin{table}[t]
\renewcommand{\arraystretch}{1}
\centering
\vspace{0.1cm}
\caption{Ablation study on collision evaluation.}
\label{tab:collision_ablation}
\resizebox{\linewidth}{!}{%
\begin{tabular}{|@{\hskip 4pt}c@{\hskip 4pt}|@{\hskip 4pt}c@{\hskip 4pt}|@{\hskip 4pt}c@{\hskip 4pt}|@{\hskip 4pt}c@{\hskip 4pt}|@{\hskip 4pt}c@{\hskip 4pt}|@{\hskip 4pt}c@{\hskip 4pt}|@{\hskip 4pt}c@{\hskip 4pt}|@{\hskip 4pt}c@{\hskip 4pt}|@{\hskip 4pt}c@{\hskip 4pt}|}
\hline
\textbf{Prompt} & \textbf{Method} & \textbf{TP} & \textbf{FP} & \textbf{FN} & \textbf{TN} & \textbf{Precision} & \textbf{Recall} & \textbf{F1 Score} \\
\hline
Cart & \textit{Base}        & 1 & 0 & 99 & 100 & 1.00 & 0.01 & 0.02 \\
Cart &  \textit{CP}         & 2  & 0  & 98 & 100 & 1.00 & 0.02 & 0.04 \\
Cart & \textit{CoT}         & 6  & 0  & 94 & 100 & 1.00 & 0.06 & 0.11 \\
\textbf{Cart} & \textbf{\textit{All}} & 12 & 0 & 88 & 100 & \textbf{1.00} & \textbf{0.12} & \textbf{0.21} \\
Ego & \textit{Base}         & 54 & 0 & 46 & 100 & 1.00 & 0.54 & 0.70 \\
Ego  &  \textit{CP}         & 66 & 8  & 34 & 92  & 0.89 & 0.66 & 0.79 \\
Ego  & \textit{CoT}         & 75 & 8  & 25 & 92  & 0.90 & 0.75 & 0.84 \\
\textbf{Ego} & \textbf{\textit{All}} & 96 & 20 & 4 & 80 & \textbf{0.83} & \textbf{0.96} & \textbf{0.87} \\
\hline
\end{tabular}
}
\vspace{0.1cm}
\begin{flushleft}
\scriptsize Cart and Ego denote $\mathrm{\textit{Prompt}_{\mathrm{cart}}}$ and $\mathrm{\textit{Prompt}_{\mathrm{ego}}}$, respectively. \textit{TP} (True Positive): collision correctly identified; 
\textit{FP} (False Positive): non-collision incorrectly identified as collision; 
\textit{FN} (False Negative): collision missed; 
\textit{TN} (True Negative): non-collision correctly identified.
\end{flushleft}
\end{table}

\begin{table}[b]
\renewcommand{\arraystretch}{1}
\centering
\caption{LLMs comparison for collision evaluation.}
\label{tab:llm_collision}
\resizebox{\linewidth}{!}{%
\begin{tabular}{|@{\hskip 2pt}c@{\hskip 2pt}|@{\hskip 4pt}c@{\hskip 4pt}|@{\hskip 4pt}c@{\hskip 4pt}|@{\hskip 4pt}c@{\hskip 4pt}|@{\hskip 4pt}c@{\hskip 4pt}|@{\hskip 4pt}c@{\hskip 4pt}|@{\hskip 4pt}c@{\hskip 4pt}|@{\hskip 4pt}c@{\hskip 2pt}|@{\hskip 2pt}c@{\hskip 2pt}|}
\hline
\textbf{Prompt} & \textbf{LLM} & \textbf{TP} & \textbf{FP} & \textbf{FN} & \textbf{TN} & \textbf{Precision} & \textbf{Recall} & \textbf{F1 Score} \\
\hline
\textbf{Cart} & \textbf{\textit{Gemini-1.5Pro}} & 12 & 0  & 88 & 100 & \textbf{1.00} & \textbf{0.12} & \textbf{0.21} \\
Cart & \textit{GPT-4o}     & 1 & 0  & 99 & 104 & 1.00 & 0.01 & 0.02 \\
Cart & \textit{DeepSeek-V3}   & 2 & 2  & 98 & 102 & 0.50 & 0.02 & 0.04 \\
Ego & \textit{Gemini-1.5Pro} & 96 & 20 & 4 & 80 & 0.83 & 0.96 & 0.87 \\
Ego & \textit{GPT-4o}     & 95 & 20 & 5 & 84 & 0.83 & 0.95 & 0.88 \\
\textbf{Ego} & \textbf{\textit{DeepSeek-V3}}   & 97 & 19 & 3 & 85 & \textbf{0.84} & \textbf{0.97} & \textbf{0.90} \\
\hline
\end{tabular}
}
\end{table}

As shown in Table~\ref{tab:collision_ablation}, $\mathrm{\textit{Prompt}_{\mathrm{cart}}}$ performs poorly across all settings, with a maximum F1 score of 0.21, underscoring its limitations in collision reasoning. In contrast, $\mathrm{\textit{Prompt}_{\mathrm{ego}}}$ shows clear gains with each added prompting technique, improving from an F1 score of 0.70 (\textit{Base}) to 0.87 (\textit{All}). The \textit{All} configuration achieves the highest recall (0.96), demonstrating strong detection of true collisions. While precision drops slightly from 0.89 to 0.83, this reflects a typical trade-off favoring recall in safety-critical tasks. Overall, the results confirm that CP, CoT, and ICL significantly enhance the evaluator's performance.

\subsubsection{Comparison Across Different \glspl{llm}}
We compare the reasoning performance of Gemini-1.5Pro, GPT-4o, and DeepSeek-V3 using the same $\mathrm{\textit{Prompt}_{\mathrm{cart}}}$ and $\mathrm{\textit{Prompt}_{\mathrm{ego}}}$ with \textit{All} configuration applied. As shown in Table~\ref{tab:llm_collision}, all models perform significantly better with $\mathrm{\textit{Prompt}_{\mathrm{ego}}}$, confirming the findings in Section~\ref{sec:ablation}. Gemini-1.5Pro achieves the highest F1 score (0.21) with $\mathrm{\textit{Prompt}_{\mathrm{cart}}}$, likely due to its larger context window (up to 200k tokens), which helps interpret verbose Cartesian inputs. With ego-centric prompts, DeepSeek-V3 slightly outperforms the others (F1 score 0.90), followed by GPT-4o (0.88) and Gemini-1.5Pro (0.87).%, suggesting a slight edge in reasoning ability for DeepSeek-V3.
%. This consistent trend reinforces our earlier findings on the superiority of ego-centric representations for safety-critical scenario understanding.
% \begin{figure}[t]
%     \centering
%     \includegraphics[width=0.38\textwidth]{figures/f1score_scenario.jpg}
%     \caption{F1 score comparison for collision scenario evaluation across LLMs.}
%     \label{fig: f1score_scenario}
% \end{figure}

This evaluation approach can be readily extended to general safety-criticality assessments, such as classifying scenarios as safe or high-risk, as it only requires structured scenario data. A generalized example is available in our GitHub repository. Since traditional metric-based methods are used to generate the ground truth, we do not perform direct comparisons against them.
\subsection{Safety Inference in Driving Scenarios}
\begin{figure}[t]
    \centering
    \includegraphics[width=0.495\textwidth]{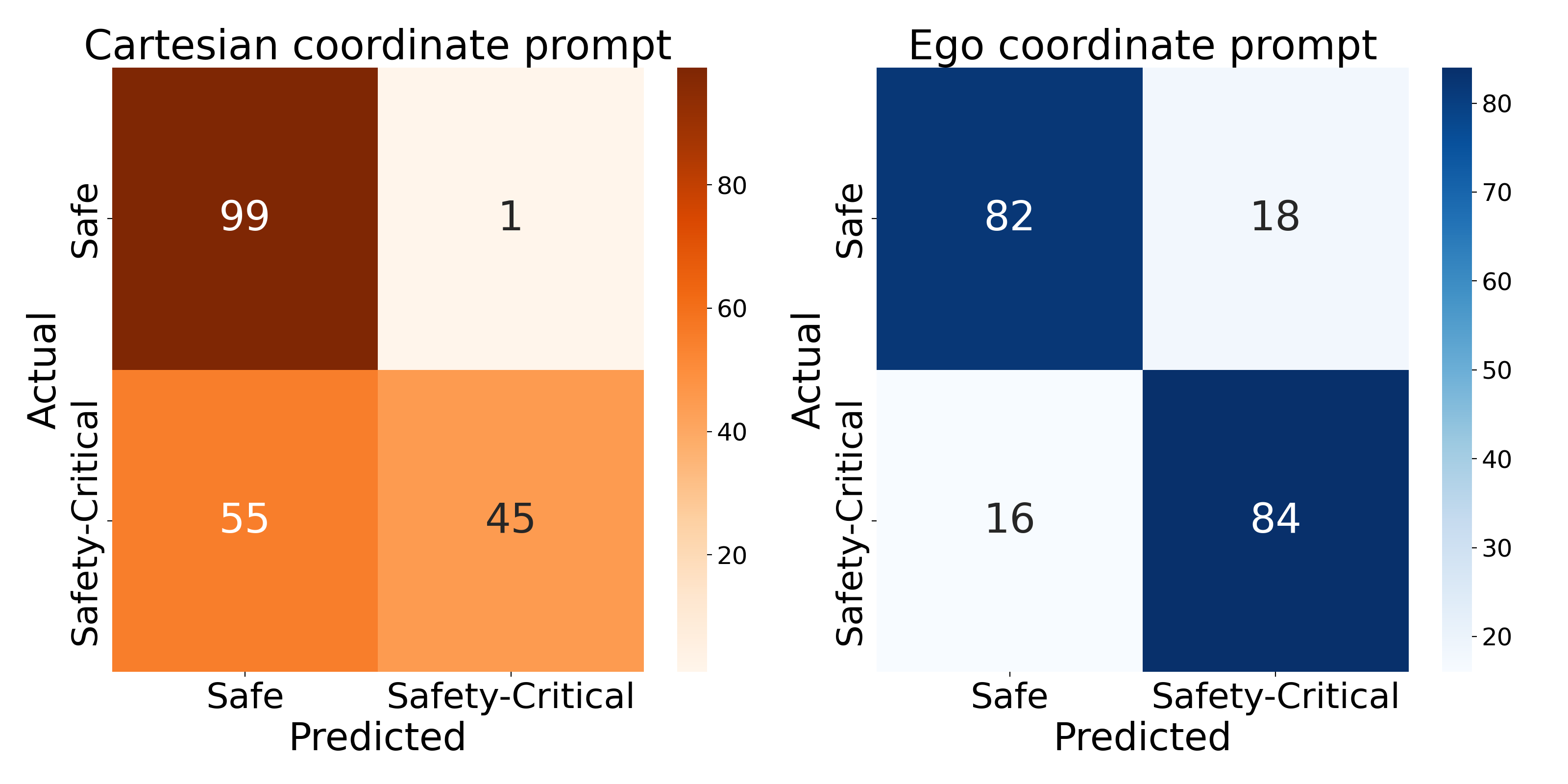}
    \caption{Confusion matrices for safety evaluation, using $\mathrm{\textit{Prompt}_{\mathrm{cart}}}$ and $\mathrm{\textit{Prompt}_{\mathrm{ego}}}$.}
    \label{fig: confusion_agents}
\end{figure}

Another application of our framework is safety inference, which aims to predict whether a scenario will become safety-critical for the \gls{ev} in the near future. This prediction is based on the current state and a 10-timestep (\SI{1}{\second}) history of the environment, considering all surrounding vehicles within a \SI{30}{\meter} radius.
We evaluate Gemini-1.5Pro using both $\mathrm{\textit{Prompt}_{\mathrm{cart}}}$ and $\mathrm{\textit{Prompt}_{\mathrm{ego}}}$, combined with the safety-criticality prediction output format described in Section~\ref{sec:output}. As shown in Fig.~\ref{fig: confusion_agents}, $\mathrm{\textit{Prompt}_{\mathrm{cart}}}$ performs well on safe cases (99 correctly predicted) but struggles with safety-critical ones, correctly identifying only 45 of them. This illustrates the limitations of Cartesian-based inputs for forward-looking risk reasoning. In contrast, $\mathrm{\textit{Prompt}_{\mathrm{ego}}}$ significantly improves performance, correctly predicting 84 safety-critical and 82 safe cases. These results confirm the value of ego-centric inputs, which more effectively capture dynamic interactions from the \gls{ev}'s perspective, enabling more accurate safety inference.

\subsubsection{Ablation Study}
We conduct an ablation study using four configurations \textit{Base}, \textit{CP}, \textit{CoT}, and \textit{All} applied to both $\mathrm{\textit{Prompt}_{\mathrm{cart}}}$ and $\mathrm{\textit{Prompt}_{\mathrm{ego}}}$ to assess their impact on safety inference. Table~\ref{tab:ablation_safety_eval} shows that $\mathrm{\textit{Prompt}_{\mathrm{ego}}}$ consistently outperforms $\mathrm{\textit{Prompt}_{\mathrm{cart}}}$. Notably, $\mathrm{\textit{Prompt}_{\mathrm{cart}}}$ achieves solid performance in the safety inference task, with the \textit{All} configuration reaching an F1 score of 0.62. This result suggests that Cartesian inputs can be effective when the task is limited to nearby agents and local context.

\begin{table}[b]
\renewcommand{\arraystretch}{1}
\centering
\caption{Ablation study on safety inference.}
\label{tab:ablation_safety_eval}
\resizebox{\linewidth}{!}{%
\begin{tabular}{|@{\hskip 4pt}c@{\hskip 4pt}|@{\hskip 4pt}c@{\hskip 4pt}|@{\hskip 4pt}c@{\hskip 4pt}|@{\hskip 4pt}c@{\hskip 4pt}|@{\hskip 4pt}c@{\hskip 4pt}|@{\hskip 4pt}c@{\hskip 4pt}|@{\hskip 4pt}c@{\hskip 4pt}|@{\hskip 4pt}c@{\hskip 4pt}|@{\hskip 4pt}c@{\hskip 4pt}|}
\hline
\textbf{Prompt} & \textbf{Method} & \textbf{TP} & \textbf{FP} & \textbf{FN} & \textbf{TN} & \textbf{Precision} & \textbf{Recall} & \textbf{F1 Score} \\
\hline
Cart & \textit{Base} & 4 & 18 & 96 & 82 & 0.18 & 0.04 & 0.07 \\
Cart & \textit{CP}  & 11 & 6  & 89 & 94  & 0.65 & 0.11 & 0.19 \\
Cart & \textit{CoT}  & 42 & 8  & 58 & 92  & 0.84 & 0.42 & 0.56 \\
\textbf{Cart} & \textbf{\textit{All}}  & 45 & 1  & 55 & 99 & \textbf{0.98} & \textbf{0.45} & \textbf{0.62} \\
Ego  & \textit{Base} & 64 & 23 & 36 & 77 & 0.74 & 0.64 & 0.69 \\
Ego  & \textit{CP}  & 65 & 24 & 35 & 76  & 0.72 & 0.65 & 0.68 \\
Ego  & \textit{CoT}  & 73 & 16 & 27 & 84  & 0.82 & 0.73 & 0.77 \\
\textbf{Ego} & \textbf{\textit{All}}  & 84 & 18 & 16 & 82 & \textbf{0.82} & \textbf{0.84} & \textbf{0.83} \\
\hline
\end{tabular}
}
\end{table}
\begin{table}[h]
\renewcommand{\arraystretch}{1}
\centering
\caption{LLMs comparison for safety inference.}
\label{tab:llm_safety_eval}
\resizebox{\linewidth}{!}{%
\begin{tabular}{|@{\hskip 2pt}c@{\hskip 2pt}|@{\hskip 4pt}c@{\hskip 4pt}|@{\hskip 4pt}c@{\hskip 4pt}|@{\hskip 4pt}c@{\hskip 4pt}|@{\hskip 4pt}c@{\hskip 4pt}|@{\hskip 4pt}c@{\hskip 4pt}|@{\hskip 4pt}c@{\hskip 4pt}|@{\hskip 4pt}c@{\hskip 2pt}|@{\hskip 2pt}c@{\hskip 2pt}|}
\hline
\textbf{Prompt} & \textbf{LLM} & \textbf{TP} & \textbf{FP} & \textbf{FN} & \textbf{TN} & \textbf{Precision} & \textbf{Recall} & \textbf{F1 Score} \\
\hline
\textbf{Cart} & \textbf{\textit{Gemini-1.5Pro}}     & 45 & 1  & 55 & 99 & \textbf{0.98} & \textbf{0.45} & \textbf{0.62} \\
Cart & \textit{GPT-4o}     & 22 & 9  & 78 & 91 & 0.71 & 0.22 & 0.34 \\
Cart & \textit{\textit{DeepSeek-V3}}   & 37 & 2  & 63 & 98 & 0.95 & 0.37 & 0.53 \\
Ego       & \textit{Gemini-1.5Pro}     & 84 & 18 & 16 & 82 & 0.82 & 0.84 & 0.83 \\
Ego       & \textit{GPT-4o}     & 85 & 24 & 15 & 76 & 0.78 & 0.85 & 0.81 \\
\textbf{Ego}       & \textbf{\textit{DeepSeek-V3}}   & 89 & 16 & 11 & 84 & \textbf{0.85} & \textbf{0.89} & \textbf{0.87} \\
\hline
\end{tabular}
}
\end{table}

However, the $\mathrm{\textit{Prompt}_{\mathrm{ego}}}$ yields strong and balanced results across all settings, with the \textit{All} configuration achieving the highest F1 score (0.83) and recall (0.84). These findings reinforce that ego-centric prompts, especially when enhanced by CP, ICL, and CoT techniques, are most effective for safety inference.

\subsubsection{Comparison Across Different \glspl{llm}}
To assess the reasoning performance of different \glspl{llm} in the safety inference task, we apply the same $\mathrm{\textit{Prompt}_{\mathrm{cart}}}$ and $\mathrm{\textit{Prompt}_{\mathrm{ego}}}$ using the \textit{All} prompting configuration.  The results are summarized in Table~\ref{tab:llm_safety_eval}. All models perform significantly better with $\mathrm{\textit{Prompt}_{\mathrm{ego}}}$, reaffirming the value of ego-centric representations. DeepSeek-V3 achieves the highest F1 score (0.87), followed by Gemini-1.5Pro (0.83) and GPT-4o (0.81). On Cartesian input, Gemini-1.5Pro leads with an F1 score of 0.62, outperforming GPT-4o (0.34) and DeepSeek-V3 (0.53), likely due to its larger context window, which helps interpret verbose global inputs. In contrast, DeepSeek-V3 slightly outperforms the others in terms of ego-centric input. This suggests that model reasoning performance plays a greater role when context length becomes less critical.

Overall, our safety inference framework operates independently during simulation and can serve as a supplementary module to support motion planners in online risk assessment. While the LLM-based evaluator shows strong reasoning capabilities, its current response time, averaging several seconds per inference, remains unsuitable for real-time deployment. Future work may explore fine-tuning smaller, task-specific models to enable faster, safer inference. Additional runtime details and performance metrics are provided in our GitHub repository.
\begin{figure}[b]
    \centering
    \input{figures/Riskscore}
    \caption{Risk score of the ego-attacker vehicle's trajectory, in the original safe scenario and in the generated safety-critical one. A risk score of 5 indicates very low risk, while 0 denotes a collision.}
    \label{fig: risk_score}
\end{figure}
\subsection{Generation of Safety-Critical Scenarios}
Our safety-critical scenario generation framework identifies an ego-attacker and adversarially modifies its trajectory. Using the safety-critical metrics prompt (Sec~\ref{sec:prompt_generation}), the ego-attacker would be determined based on these risk assessments and motion description within \SI{30}{\meter} in front or to the side of the \gls{ev}. Then, a trajectory generation prompt synthesizes candidate scenarios using Self-Consistency (SC) and selects the most adversarial outcome,i.e., the trajectory yielding the lowest risk score for the identified ego-attacker.
\begin{figure}[t]
    \centering
    \includegraphics[width=0.45\textwidth]{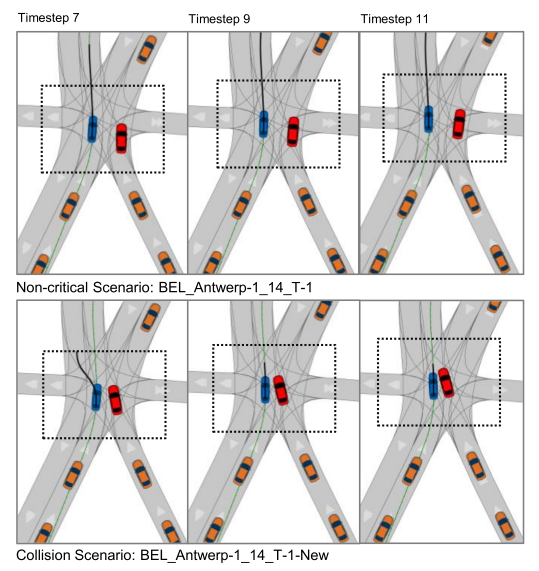}
    \caption{
        A case study of safety-critical scenario generation on CommonRoad\cite{althoff2017commonroad} BEL Antwerp-1\_14\_T-1. The ego vehicle is in blue, and the ego-attacker vehicle is in red. The top row depicts the original safe scenario, while the bottom row shows the corresponding safety-critical scenario generated by our model.
    }
    \label{fig: Scenario_generation}
\end{figure}
To assess the effectiveness of our framework, we conduct a case study using a scenario from the previous 4,700 safe scenarios from Sec~\ref{sec:setup}. As shown in Fig.~\ref{fig: risk_score} and \ref{fig: Scenario_generation}, the original scenario is safe, and the ego-attacker's trajectory has a safe risk level. Conversely, in the generated scenario, the ego-attacker's modified trajectory has a decreasing risk score over time (Fig.~\ref{fig: risk_score}), ultimately leading to a collision.
Fig.~\ref{fig: Scenario_generation} further visualizes this evolution. The top row shows the original non-critical scenario, where the blue \gls{ev} and the red ego-attacker navigate safely. The bottom row illustrates the modified scenario, where a collision is triggered. These results validate the ability of our framework to generate targeted, safety-critical scenarios based on structured LLM-driven risk inference.

Since we focus on identifying the ego-attacker and generating safety-critical scenarios adversarially, our framework can provide greater efficiency and controllability compared to purely text-based scenario generation approaches integrating \glspl{llm}. 

\section{Conclusion \& Outlook}
\label{sec:conclusion}
This paper introduced a novel \gls{llm}-based framework for evaluating and generating safety-critical driving scenarios from structured data. We proposed two prompt templates $\mathrm{\textit{Prompt}_{\mathrm{cart}}}$ and $\mathrm{\textit{Prompt}_{\mathrm{ego}}}$ supported by parsers, and advanced prompting techniques including Contextual Prompting, Chain-of-Thought reasoning, and In-Context Learning. We also developed an LLM-guided scenario generation module to generate
safety-critical scenarios by adversarially modifying
ego-attackers' trajectories to create the safety-critical cases.

The framework was evaluated in three use cases supported by structured outputs: collision scenario evaluation, safety inference, and safety-critical scenario generation. Across all \glspl{llm}, $\mathrm{\textit{Prompt}_{\mathrm{ego}}}$ consistently outperformed $\mathrm{\textit{Prompt}_{\mathrm{cart}}}$, for the Gemini-1.5Pro, achieving F1 scores of 0.87 vs. 0.21 in collision evaluation, and 0.83 vs. 0.62 in safety inference, highlighting the advantage of ego-centric representations for reasoning about risk. In generation, the framework successfully minimized the ego-attacker’s risk score to synthesize safety-critical outcomes. Future work includes extending the framework to 3D simulators such as CARLA, enhancing the generation module to produce more diverse safety-critical scenarios, and using the generated scenarios to further evaluate motion planning algorithms \cite{Piazza2024}.
%%%%%%%%%%%%%%%%%%%%%%%%%%%%%%%%%%%%%%%%%%%%%%%%%%%%%%%%%
%%% Bibliography
%%%%%%%%%%%%%%%%%%%%%%%%%%%%%%%%%%%%%%%%%%%%%%%%%%%%%%%%%
\bibliographystyle{IEEEtran}
\bibliography{literatur}
\end{document}

%% file: figures/Riskscore.tex
\definecolor{TUMBlue}{HTML}{0065bd}
\definecolor{TUMOrange}{HTML}{E37222}

\begin{tikzpicture}[font=\footnotesize]

\begin{axis}[
yshift=-0.1cm,
anchor=north,
/pgf/number format/.cd,
1000 sep={},
height=4.5cm,
width=6.5cm,
legend style={
	at={(0.64,0.98)}, 
	anchor=north,
	legend columns=1,
	cells={anchor=west},
	draw=black,
	column sep=0.25em,
	row sep=0.05em,
    font=\scriptsize
},
scale only axis,
scaled ticks=false,
scaled ticks=false,
tick label style={/pgf/number format/fixed},
xlabel={Timestep},
ylabel={Risk Score (0-5)},
x label style={at={(0.5,-0.07)},anchor=north},
xmin=0, xmax=15,
ymin=-0.5, ymax=5.5,
ytick = {0, 1, ..., 5},
ymajorgrids=true,
xmajorgrids=true,
]
\addplot [ultra thick, TUMBlue, solid, mark=*, mark size=1.5pt]
table {%
1 3
2 3
3 3
4 4
5 4
6 4
7 4
8 4
9 4
10 4
11 4
12 4
13 4
14 4
};
\addlegendentry{Original safe scenario}

\addplot [ultra thick, TUMOrange, solid, mark=square*, mark size=1.5pt]
table {%
1 3
2 3
3 3
4 2
5 2
6 2
7 2
8 1
9 1
10 1
11 0
};
\addlegendentry{Generated safety-critical scenario}

\node[inner sep=0pt] at (axis cs:11, 0) {\includegraphics[height=6mm]{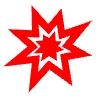}};

\end{axis}

\end{tikzpicture}

%% file: main.bbl
% Generated by IEEEtran.bst, version: 1.14 (2015/08/26)
\begin{thebibliography}{10}
\providecommand{\url}[1]{#1}
\csname url@samestyle\endcsname
\providecommand{\newblock}{\relax}
\providecommand{\bibinfo}[2]{#2}
\providecommand{\BIBentrySTDinterwordspacing}{\spaceskip=0pt\relax}
\providecommand{\BIBentryALTinterwordstretchfactor}{4}
\providecommand{\BIBentryALTinterwordspacing}{\spaceskip=\fontdimen2\font plus
\BIBentryALTinterwordstretchfactor\fontdimen3\font minus
  \fontdimen4\font\relax}
\providecommand{\BIBforeignlanguage}[2]{{%
\expandafter\ifx\csname l@#1\endcsname\relax
\typeout{** WARNING: IEEEtran.bst: No hyphenation pattern has been}%
\typeout{** loaded for the language `#1'. Using the pattern for}%
\typeout{** the default language instead.}%
\else
\language=\csname l@#1\endcsname
\fi
#2}}
\providecommand{\BIBdecl}{\relax}
\BIBdecl

\bibitem{Betz2024}
J.~Betz, M.~Lutwitzi, and S.~Peters, ``A new taxonomy for automated driving:
  Structuring applications based on their operational design domain, level of
  automation and automation readiness,'' in \emph{2024 IEEE Intelligent
  Vehicles Symposium (IV)}, 2024, pp. 1--7.

\bibitem{khan2023safety}
F.~Khan, M.~Falco, H.~Anwar, and D.~Pfahl, ``Safety testing of automated
  driving systems: A literature review,'' \emph{IEEE Access}, 2023.

\bibitem{riedmaier2020survey}
S.~Riedmaier, T.~Ponn, D.~Ludwig, B.~Schick, and F.~Diermeyer, ``Survey on
  scenario-based safety assessment of automated vehicles,'' \emph{IEEE access},
  vol.~8, pp. 87\,456--87\,477, 2020.

\bibitem{hallerbach2018simulation}
S.~Hallerbach, Y.~Xia, U.~Eberle, and F.~Koester, ``Simulation-based
  identification of critical scenarios for cooperative and automated
  vehicles,'' \emph{SAE International Journal of Connected and Automated
  Vehicles}, vol.~1, no. 2018-01-1066, pp. 93--106, 2018.

\bibitem{brown2020language}
T.~Brown, B.~Mann, N.~Ryder, M.~Subbiah, J.~D. Kaplan, P.~Dhariwal,
  A.~Neelakantan, P.~Shyam, G.~Sastry, A.~Askell \emph{et~al.}, ``Language
  models are few-shot learners,'' \emph{Advances in neural information
  processing systems}, vol.~33, pp. 1877--1901, 2020.

\bibitem{vaswani2017attention}
A.~Vaswani, ``Attention is all you need,'' \emph{Advances in Neural Information
  Processing Systems}, 2017.

\bibitem{sahoo2024systematic}
P.~Sahoo, A.~K. Singh, S.~Saha, V.~Jain, S.~Mondal, and A.~Chadha, ``A
  systematic survey of prompt engineering in large language models: Techniques
  and applications,'' \emph{arXiv preprint arXiv:2402.07927}, 2024.

\bibitem{wu2023language}
D.~Wu, W.~Han, T.~Wang, Y.~Liu, X.~Zhang, and J.~Shen, ``Language prompt for
  autonomous driving,'' \emph{arXiv preprint arXiv:2309.04379}, 2023.

\bibitem{wang2024dualad}
D.~Wang, M.~Kaufeld, and J.~Betz, ``Dualad: Dual-layer planning for reasoning
  in autonomous driving,'' \emph{arXiv preprint arXiv:2409.18053}, 2024.

\bibitem{xu2024drivegpt4}
Z.~Xu, Y.~Zhang, E.~Xie, Z.~Zhao, Y.~Guo, K.-Y.~K. Wong, Z.~Li, and H.~Zhao,
  ``Drivegpt4: Interpretable end-to-end autonomous driving via large language
  model,'' \emph{IEEE Robotics and Automation Letters}, 2024.

\bibitem{song2023critical}
Q.~Song, K.~Tan, P.~Runeson, and S.~Persson, ``Critical scenario identification
  for realistic testing of autonomous driving systems,'' \emph{Software Quality
  Journal}, vol.~31, no.~2, pp. 441--469, 2023.

\bibitem{wu2024reality}
J.~Wu, C.~Lu, A.~Arrieta, T.~Yue, and S.~Ali, ``Reality bites: Assessing the
  realism of driving scenarios with large language models,'' in
  \emph{Proceedings of the 2024 IEEE/ACM First International Conference on AI
  Foundation Models and Software Engineering}, 2024, pp. 40--51.

\bibitem{lu2023deepscenario}
C.~Lu, T.~Yue, and S.~Ali, ``Deepscenario: An open driving scenario dataset for
  autonomous driving system testing,'' in \emph{2023 IEEE/ACM 20th
  International Conference on Mining Software Repositories (MSR)}.\hskip 1em
  plus 0.5em minus 0.4em\relax IEEE, 2023, pp. 52--56.

\bibitem{you2025comprehensive}
S.~You, X.~Luo, X.~Liang, J.~Yu, C.~Zheng, and J.~Gong, ``A comprehensive
  llm-powered framework for driving intelligence evaluation,'' \emph{arXiv
  preprint arXiv:2503.05164}, 2025.

\bibitem{Dosovitskiy17}
A.~Dosovitskiy, G.~Ros, F.~Codevilla, A.~Lopez, and V.~Koltun, ``{CARLA}: {An}
  open urban driving simulator,'' in \emph{Proceedings of the 1st Annual
  Conference on Robot Learning}, 2017, pp. 1--16.

\bibitem{lu2024multimodal}
Q.~Lu, X.~Wang, Y.~Jiang, G.~Zhao, M.~Ma, and S.~Feng, ``Multimodal large
  language model driven scenario testing for autonomous vehicles,'' \emph{arXiv
  preprint arXiv:2409.06450}, 2024.

\bibitem{gao2025foundation}
Y.~Gao, M.~Piccinini, Y.~Zhang, D.~Wang, K.~Moller, R.~Brusnicki, B.~Zarrouki,
  A.~Gambi, J.~F. Totz, K.~Storms \emph{et~al.}, ``Foundation models in
  autonomous driving: A survey on scenario generation and scenario analysis,''
  \emph{IEEE Open Journal of Intelligent Transportation Systems}, 2025,
  submitted.

\bibitem{ding2023survey}
W.~Ding, C.~Xu, M.~Arief, H.~Lin, B.~Li, and D.~Zhao, ``A survey on
  safety-critical driving scenario generation—a methodological perspective,''
  \emph{IEEE Transactions on Intelligent Transportation Systems}, vol.~24,
  no.~7, pp. 6971--6988, 2023.

\bibitem{wheeler2016factor}
T.~A. Wheeler and M.~J. Kochenderfer, ``Factor graph scene distributions for
  automotive safety analysis,'' in \emph{2016 IEEE 19th International
  Conference on Intelligent Transportation Systems (ITSC)}.\hskip 1em plus
  0.5em minus 0.4em\relax IEEE, 2016, pp. 1035--1040.

\bibitem{ding2018new}
W.~Ding, W.~Wang, and D.~Zhao, ``A new multi-vehicle trajectory generator to
  simulate vehicle-to-vehicle encounters,'' \emph{arXiv preprint
  arXiv:1809.05680}, 2018.

\bibitem{jain2019analyzing}
L.~Jain, V.~Chandrasekaran, U.~Jang, W.~Wu, A.~Lee, A.~Yan, S.~Chen, S.~Jha,
  and S.~A. Seshia, ``Analyzing and improving neural networks by generating
  semantic counterexamples through differentiable rendering,'' \emph{arXiv
  preprint arXiv:1910.00727}, 2019.

\bibitem{sun2021corner}
H.~Sun, S.~Feng, X.~Yan, and H.~X. Liu, ``Corner case generation and analysis
  for safety assessment of autonomous vehicles,'' \emph{Transportation research
  record}, vol. 2675, no.~11, pp. 587--600, 2021.

\bibitem{rana2021building}
A.~Rana and A.~Malhi, ``Building safer autonomous agents by leveraging risky
  driving behavior knowledge,'' in \emph{2021 International Conference on
  Communications, Computing, Cybersecurity, and Informatics (CCCI)}.\hskip 1em
  plus 0.5em minus 0.4em\relax IEEE, 2021, pp. 1--6.

\bibitem{cao2023robust}
Y.~Cao, D.~Xu, X.~Weng, Z.~Mao, A.~Anandkumar, C.~Xiao, and M.~Pavone, ``Robust
  trajectory prediction against adversarial attacks,'' in \emph{Conference on
  Robot Learning}.\hskip 1em plus 0.5em minus 0.4em\relax PMLR, 2023, pp.
  128--137.

\bibitem{SUMO2018}
\BIBentryALTinterwordspacing
P.~A. Lopez, M.~Behrisch, L.~Bieker-Walz, J.~Erdmann, Y.-P. Fl{\"o}tter{\"o}d,
  R.~Hilbrich, L.~L{\"u}cken, J.~Rummel, P.~Wagner, and E.~Wie{\ss}ner,
  ``Microscopic traffic simulation using sumo,'' in \emph{The 21st IEEE
  International Conference on Intelligent Transportation Systems}.\hskip 1em
  plus 0.5em minus 0.4em\relax IEEE, 2018. [Online]. Available:
  \url{https://elib.dlr.de/124092/}
\BIBentrySTDinterwordspacing

\bibitem{zhang2024chatscene}
J.~Zhang, C.~Xu, and B.~Li, ``Chatscene: Knowledge-enabled safety-critical
  scenario generation for autonomous vehicles,'' in \emph{Proceedings of the
  IEEE/CVF Conference on Computer Vision and Pattern Recognition}, 2024, pp.
  15\,459--15\,469.

\bibitem{ruan2024traffic}
B.-K. Ruan, H.-T. Tsui, Y.-H. Li, and H.-H. Shuai, ``Traffic scene generation
  from natural language description for autonomous vehicles with large language
  model,'' \emph{arXiv preprint arXiv:2409.09575}, 2024.

\bibitem{li2024chatsumo}
S.~Li, T.~Azfar, and R.~Ke, ``Chatsumo: Large language model for automating
  traffic scenario generation in simulation of urban mobility,'' \emph{IEEE
  Transactions on Intelligent Vehicles}, 2024.

\bibitem{althoff2017commonroad}
M.~Althoff, M.~Koschi, and S.~Manzinger, ``Commonroad: Composable benchmarks
  for motion planning on roads,'' in \emph{2017 IEEE Intelligent Vehicles
  Symposium (IV)}.\hskip 1em plus 0.5em minus 0.4em\relax IEEE, 2017, pp.
  719--726.

\bibitem{bender2014lanelets}
P.~Bender, J.~Ziegler, and C.~Stiller, ``Lanelets: Efficient map representation
  for autonomous driving,'' in \emph{2014 IEEE Intelligent Vehicles Symposium
  Proceedings}.\hskip 1em plus 0.5em minus 0.4em\relax IEEE, 2014, pp.
  420--425.

\bibitem{Frenetix}
R.~Trauth, K.~Moller, G.~Würsching, and J.~Betz, ``Frenetix: A
  high-performance and modular motion planning framework for autonomous
  driving,'' \emph{IEEE Access}, pp. 1--1, 2024.

\bibitem{achiam2023gpt}
J.~Achiam, S.~Adler, S.~Agarwal, L.~Ahmad, I.~Akkaya, F.~L. Aleman, D.~Almeida,
  J.~Altenschmidt, S.~Altman, S.~Anadkat \emph{et~al.}, ``Gpt-4 technical
  report,'' \emph{arXiv preprint arXiv:2303.08774}, 2023.

\bibitem{jiang2023structgpt}
J.~Jiang, K.~Zhou, Z.~Dong, K.~Ye, W.~X. Zhao, and J.-R. Wen, ``Structgpt: A
  general framework for large language model to reason over structured data,''
  \emph{arXiv preprint arXiv:2305.09645}, 2023.

\bibitem{team2023gemini}
G.~Team, R.~Anil, S.~Borgeaud, J.-B. Alayrac, J.~Yu, R.~Soricut, J.~Schalkwyk,
  A.~M. Dai, A.~Hauth, K.~Millican \emph{et~al.}, ``Gemini: a family of highly
  capable multimodal models,'' \emph{arXiv preprint arXiv:2312.11805}, 2023.

\bibitem{bi2024deepseek}
X.~Bi, D.~Chen, G.~Chen, S.~Chen, D.~Dai, C.~Deng, H.~Ding, K.~Dong, Q.~Du,
  Z.~Fu \emph{et~al.}, ``Deepseek llm: Scaling open-source language models with
  longtermism,'' \emph{arXiv preprint arXiv:2401.02954}, 2024.

\bibitem{Piazza2024}
M.~Piazza, M.~Piccinini, S.~Taddei, and F.~Biral, ``Mptree: A sampling-based
  vehicle motion planner for real-time obstacle avoidance,''
  \emph{IFAC-PapersOnLine}, vol.~58, no.~10, pp. 146--153, 2024, 17th IFAC
  Symposium on Control of Transportation Systems CTS 2024.

\end{thebibliography}
